\documentclass[journal]{IEEEtran}
\usepackage{graphicx}
\usepackage{subfigure}
\usepackage{tabularx}
\usepackage{booktabs}
\usepackage{array}
\usepackage{multirow}
\usepackage{xcolor}
\newcommand{\PreserveBackslash}[1]{\let\temp=\\#1\let\\=\temp}
\newcolumntype{C}[1]{>{\PreserveBackslash\centering}p{#1}}
\newcolumntype{R}[1]{>{\PreserveBackslash\raggedleft}p{#1}}
\newcolumntype{L}[1]{>{\PreserveBackslash\raggedright}p{#1}}

\ifCLASSINFOpdf
\else
\fi
\begin{document}

\title{Training and Testing Object Detectors with \\Virtual Images}
%修改部分标记红色，若不标记，则用\newcommand{\mycolor}[0]{black}
\newcommand{\mycolor}[0]{black}

\author{Yonglin~Tian,~Xuan~Li,~Kunfeng~Wang,~\IEEEmembership{Member,~IEEE},
        and~Fei-Yue~Wang,~\IEEEmembership{Fellow,~IEEE}% <-this % stops a space
\thanks{This work was supported by National Natural Science Foundation of China (61533019, 71232006).}
% <-this % stops a space
\thanks{Yonglin Tian is with the Department of Automation, University of Science
and Technology of China, Hefei 230027, China, and also with The State Key
Laboratory of Management and Control for Complex Systems, Institute of
Automation, Chinese Academy of Sciences, Beijing 100190, China (e-mail: tyldyx@mail.ustc.edu.cn).}
\thanks{Xuan Li is with the School of Automation, Beijing Institute of Technology,
Beijing 100081, China (e-mail: lixuan0125@126.com).}% <-this % stops a space
\thanks{Kunfeng Wang (Corresponding author) is with The State Key Laboratory of Management and Control for Complex Systems, Institute of Automation, Chinese Academy of Sciences, Beijing 100190, China, and
also with Qingdao Academy of Intelligent Industries, Qingdao 266000, China
(e-mail: kunfeng.wang@ia.ac.cn).}
\thanks{Fei-Yue Wang is with The State Key Laboratory of Management and Control for
Complex Systems, Institute of Automation, Chinese Academy of Sciences, Beijing 100190, China, and also with the Research Center for Computational Experiments and Parallel Systems Technology, National University of Defense Technology, Changsha
410073, China (e-mail: feiyue@ieee.org).}% <-this % stops a space
}

\markboth{}%IEEE/CAA JOURNAL OF AUTOMATICA SINICA,~Vol.~X, No.~X, X~X
{Shell \MakeLowercase{\textit{et al.}}: Bare Demo of IEEEtran.cls
for Journals}



\maketitle

\begin{abstract}
In the area of computer vision, deep learning has produced a variety of state-of-the-art models that rely on massive labeled data. However, collecting and annotating images from the real world has a great demand for labor and money investments and is usually too passive to build datasets with specific characteristics, such as small area of objects and high occlusion level. Under the framework of Parallel Vision, this paper presents a purposeful way to design artificial scenes and automatically generate virtual images with precise annotations. A virtual dataset named ParallelEye is built, which can be used for several computer vision tasks. Then, by training the DPM (Deformable Parts Model) and Faster R-CNN detectors, we prove that the performance of models can be significantly improved by combining ParallelEye with publicly available real-world datasets during the training phase. In addition, we investigate the potential of testing the trained models from a specific aspect using intentionally designed virtual datasets, in order to discover the flaws of trained models. From the experimental results, we conclude that our virtual dataset is viable to train and test the object detectors.
\end{abstract}


% \fi
\IEEEpeerreviewmaketitle

\section{Introduction}
\IEEEPARstart{D}{atasets} play an important role in the training and testing of computer vision algorithms \cite{kaneva2011evaluation}\cite{Liu2016Visual}. However, real-world datasets are usually not satisfactory due to the insufficient diversity. And the labeling of images in real world is time-consuming and labor-intensive, especially in large-scale complex traffic systems \cite{Gou2016Vehicle}\cite{liu2017tracklet}. Moreover, it is a highly subjective work to annotate images manually. For example, different people may have different annotation results for the same image. As a result, the labeling result will deviate to some extent from the ground truth and even seriously affect the performance of computer vision algorithms.

\par
Most existing datasets originate from the real world, such as KITTI, PASCAL VOC, MS COCO, and ImageNet. Each of these datasets has many advantages, but they also have shortcomings. The KITTI \cite{geiger2012we} dataset is the world's largest computer vision dataset for automatic driving scenarios, including more than one hundred thousand labeled cars. However, KITTI lacks some common types of objects (e.g., bus), and the number of trucks is small. The PASCAL VOC \cite{everingham2010pascal} dataset serves as a benchmark for classification, recognition, and detection of visual objects. PASCAL VOC contains 20 categories, but there are small-scale images per category, with an average of less than one thousand. The ImageNet dataset \cite{deng2009imagenet} is the world's largest database of image recognition, including more than 1,000 categories. However, there is no semantic segmentation labeling information in it. There are 328,000 pictures of 91 classes of objects in the MS COCO \cite{lin2014microsoft} dataset. But the task of annotating this dataset is onerous. For example, it takes more than 20,000 hours to determine which object categories are present in the images of MS COCO.

\par
Generally speaking, real datasets are confronted with many problems, such as small scale and tedious annotation. Setting up a dataset with precise annotations from the real world means great labor and financial investments, let alone building a dataset with specific features like diverse areas of objects and occlusion levels. However, the latter occupies a significant position in addressing the problems of visual perception and understanding \cite{Wang2017}\cite{wang2016parallel}\cite{wang2017Imaging}. In work \cite{Wang2017}\cite{wang2016parallel}, Wang \emph{et al.} proposed the theoretical framework of Parallel Vision by extending the ACP approach \cite{wang2004parallel}\cite{wang2010parallel}\cite{wang2013parallel} and elaborated the significance of virtual data. The ACP methodology establishes the foundation for parallel intelligence \cite{Wang2016Where}\cite{Wang2017PDP}\cite{Wang2016Steps}, which provides a new insight to tackle issues in complex systems \cite{Li2017Parallel}. Under the framework of Parallel Vision depicted in Fig. 1, it is obvious to see the great advantage of virtual world to produce diverse labeled datasets with different environmental conditions and texture change which are usually regarded as important image features for object detection \cite{wang2015video}. In our work, we take the specifically designed virtual datasets as resources to train object detectors and also as a tool to produce feedback of the performance of trained models during the testing phase.

\par
\begin{figure}
    \centering
    \includegraphics[width=0.45\textwidth]{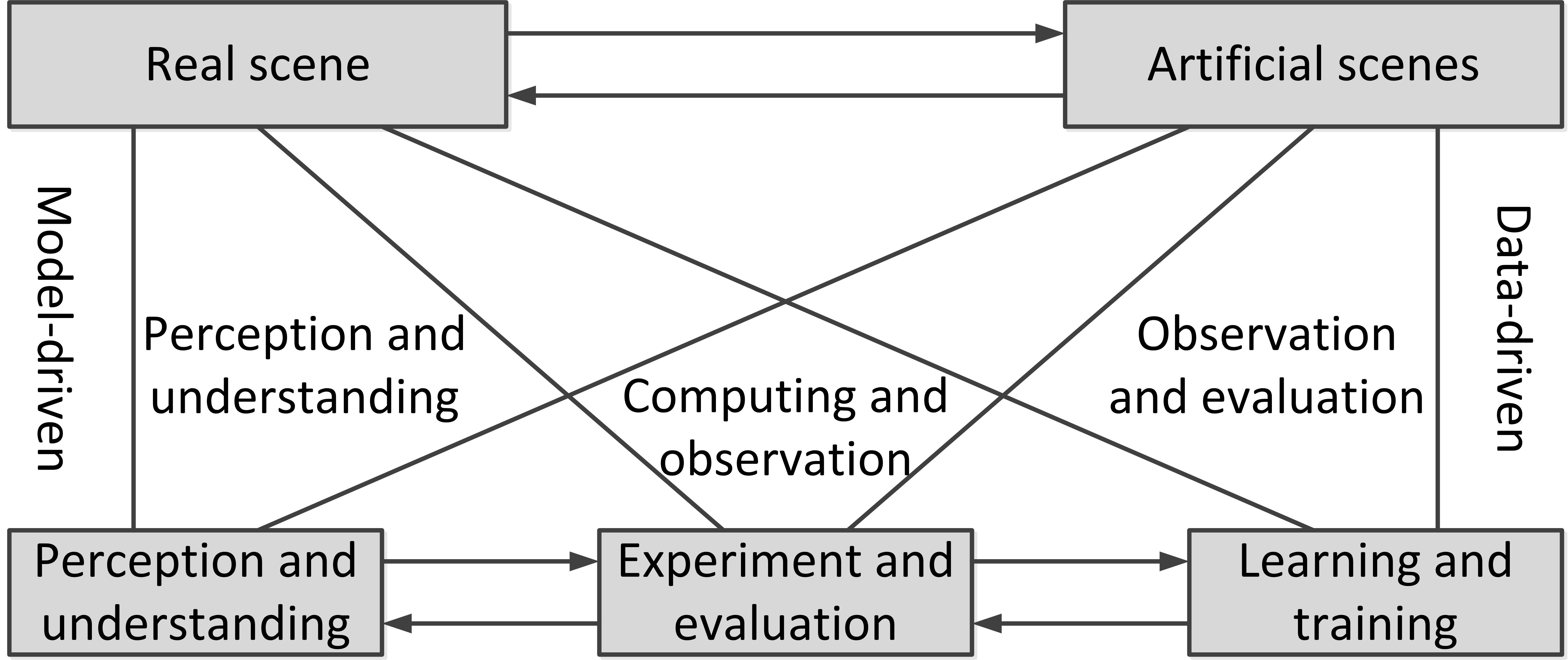}
    \caption{Basic framework and architecture for Parallel Vision.}
\end{figure}

\par
\textcolor{\mycolor}{We choose DPM (Deformable Parts Model) \cite{felzenszwalb2010object} and Faster R-CNN \cite{ren2015faster} as the object detectors in our work. DPM was one of the most effective object detectors based on HOG (Histogram of Oriented Gradient) before the resurgence of deep learning.} Faster R-CNN is currently a state of the art approach and widely used in object detection. Based on Fast R-CNN \cite{girshick2015fast}, Ren \emph{et al.} \cite{ren2015faster} introduced the Region Proposal Network (RPN) that can share convolutional features of the whole image with the detection network. This work greatly reduced the time cost to generate region proposals and improved their qualities as well. Faster R-CNN lays the foundation for many 1st-place detection models in recent years.
\par
In this paper, we present an efficient way to construct virtual image datasets with advanced computer graphics techniques. It proves a flexible and feasible method to build training datasets that can greatly satisfy our needs such as diversity, scale, and specific occlusion level. On this basis, we study the effectiveness of our virtual dataset to train and test the object detectors.
\par


%\hfill September 17, 2014

\section{Related Work}
There have been many attempts that use virtual world to carry out scientific researches. Bainbridge \emph{et al.} \cite{bainbridge2007scientific} investigated the feasibility of utilizing Second life and World of Warcraft as sites for research in the social, behavioral, and economic sciences, as well as computer science. With a virtual living lab, Prendinger \emph{et al.} \cite{prendinger2013tokyo} conducted several controlled driving and travel studies.

\par
 In the area of computer vision, early works involved training pedestrian detectors based on HOG and linear SVM \cite{marin2010learning} and part-based pedestrian detector \cite{xu2014learning} with virtual datasets generated by the video game Half-Life 2. Besides training models, virtual-world data was also used to explore the invariance of deep features of DCNNs to missing low-level cues \cite{peng2015learning} and domain adaptation issues \cite{sun2014virtual}. For semantic segmentation research, Richter \emph{et al.} \cite{richter2016playing} presented a way to build virtual datasets via modern video game and got the corresponding annotations using an outside graphics hardware without access to the source code of the game. Most of the above approaches rely on video games rather than setting up the virtual world from the scratch, resulting in bad flexibility in the research process. Recently, Ros \emph{et al.} \cite{ros2016synthia} set up a virtual world on their own and collected images with semantic annotations by virtual cameras under different weather conditions and observing angles. They generated a virtual dataset named SYNTHIA in a flexible way which was used for training DCNNs for semantic segmentations in driving scenes. However, SYNTHIA lacks the annotations for other computer vision tasks such as object detection, tracking, and so on. In a similar way, Gaidon \emph{et al.} \cite{gaidon2016virtual} proposed a real-to-virtual world cloning method and released the video dataset called ``Virtual KITTI" for multi-object tracking analysis. Basically the dataset is a clone of the real KITTI \cite{geiger2012we}, so the overall framework and layout are constricted by the real KITTI dataset. Nowdays, Generative Adversarial Networks (GANs) \cite{Goodfellow2014Generative}  are widely used to produce photorealistic synthetic images \cite{wang2017generative}, however, those images lack the corresponding annotations.

 \par
 In our work, a flexible approach to building artificial scenes from the scratch is proposed. We intend to set up virtual datasets with specific features and diverse annotations to train and test the object detectors.

\section{The Virtual Dataset ParallelEye}

\subsection{Construction of Artificial Scene}
To imitate the layout of urban scene in real world, we exported the map information of the area of interest, Zhongguancun Area, Beijing, from the platform OpenStreetMap (OSM). Then, based on the raw map, we generated buildings, roads, trees, vegetation, fences, chairs, traffic lights, traffic signs and other ``static" entities using the CGA (Computer Generated Architecture) rules of CityEngine.
Finally, we imported the scene into the game engine Unity3D where cars, buses and trucks were added. Using the C\# scripts, we controlled the vehicles to move according to certain traffic rules in the virtual world. Also, with the help of the Shaders in Unity3D, the weather and lighting condition were adjusted as needed. The artificial scene is shown in Fig. 2.
\begin{figure}
    \centering
    \includegraphics[width=0.5\textwidth]{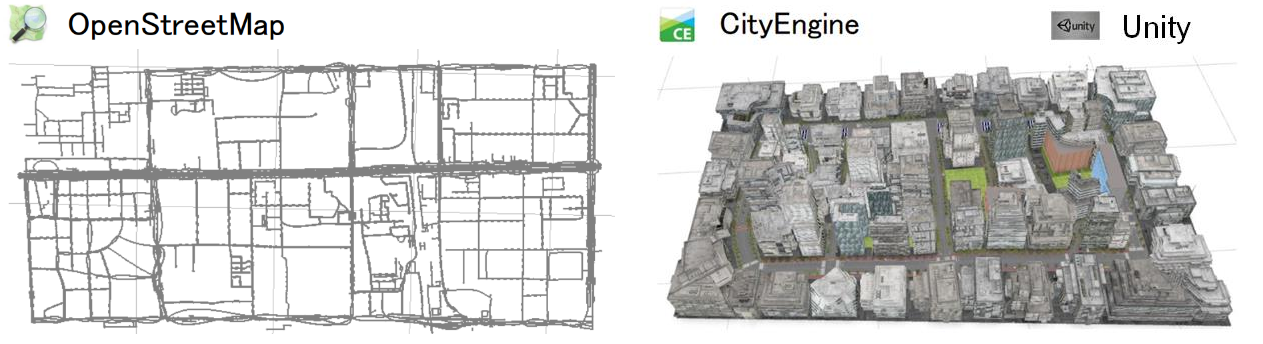}
    \caption{The appearance of artificial scene. Left: Map information exported from OSM. Right: Final artificial scene.}
\end{figure}
\subsection{Annotations of Virtual Images}
To build a dataset, we need to get the annotations of the corresponding images. Data labeling has always been a headache of the machine learning researchers. However, it is very simple and efficient to get the ground truths of the virtual images from our artificial scene via the components in Unity3D like MeshFilter, Shader, and so on. We achieved simultaneous ground truth generation including depth, optical flow, bounding box and pixel-level semantic segmentation while the scene was running. Fig. 4 shows the ground-truth annotations for different vision tasks.\par
\begin{figure}
    \centering
    \includegraphics[width=70mm]{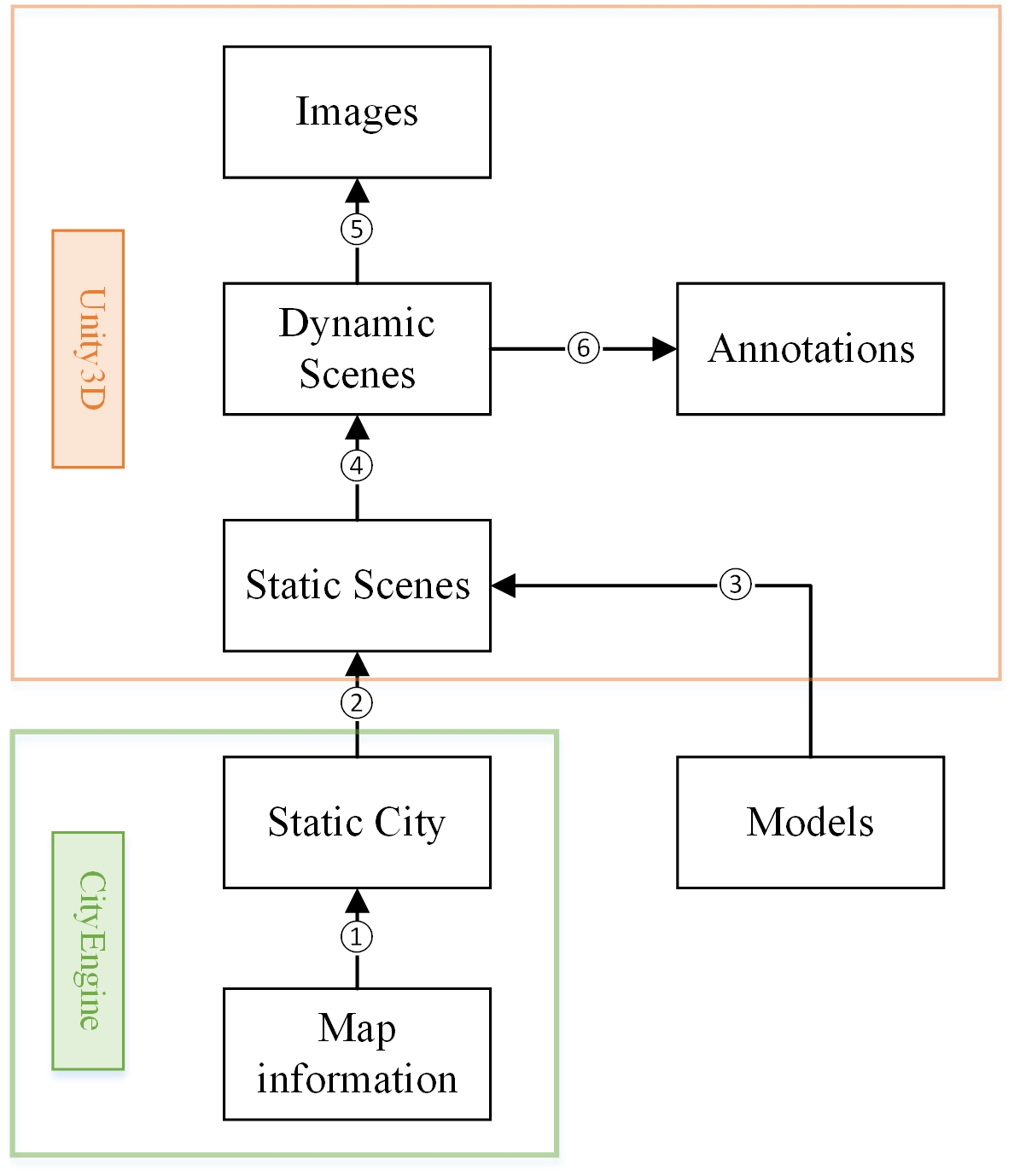}
    \caption{\textcolor{\mycolor}{Framework of constructing the virtual dataset. \textcircled{1}: In CityEngine, set up a static city including buildings and roads using the CGA rules based on the map information from OSM. \textcircled{2}\textcircled{3}: In Unity3D, import the models of interesting objects such as cars, people, animals and trees into the static city thus forming several static scenes. \textcircled{4}: In Unity3D, activate the static scenes by controlling virtual objects to move using C\# scripts. \textcircled{5}: Control the virtual camera to move and capture images using C\# scripts. \textcircled{6}: Compute the annotations such as bounding box and semantic segmentation using C\# scripts and Shaders in Unity3D.}}
\end{figure}
\begin{figure}
    \centering
    \includegraphics[width=88mm]{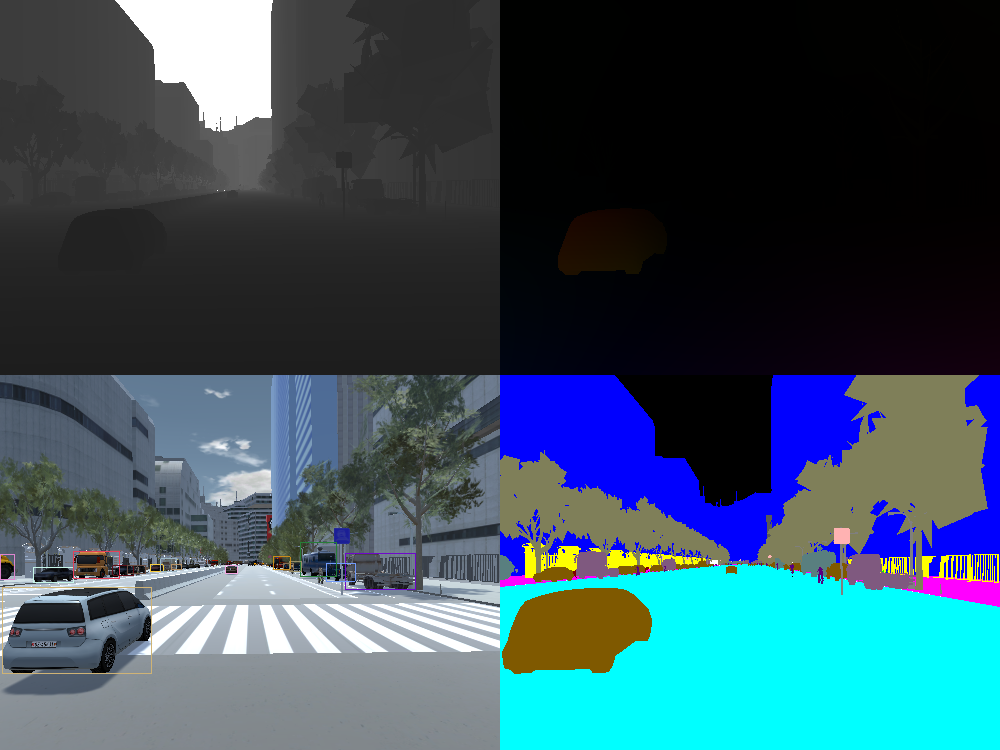}
    \caption{Annotations for different vision tasks. Top: Depth (left) and optical flow (right). Bottom: Bounding box (left) and semantic segmentation (right).}
\end{figure}
\begin{figure}
    \centering
    \includegraphics[width=88mm]{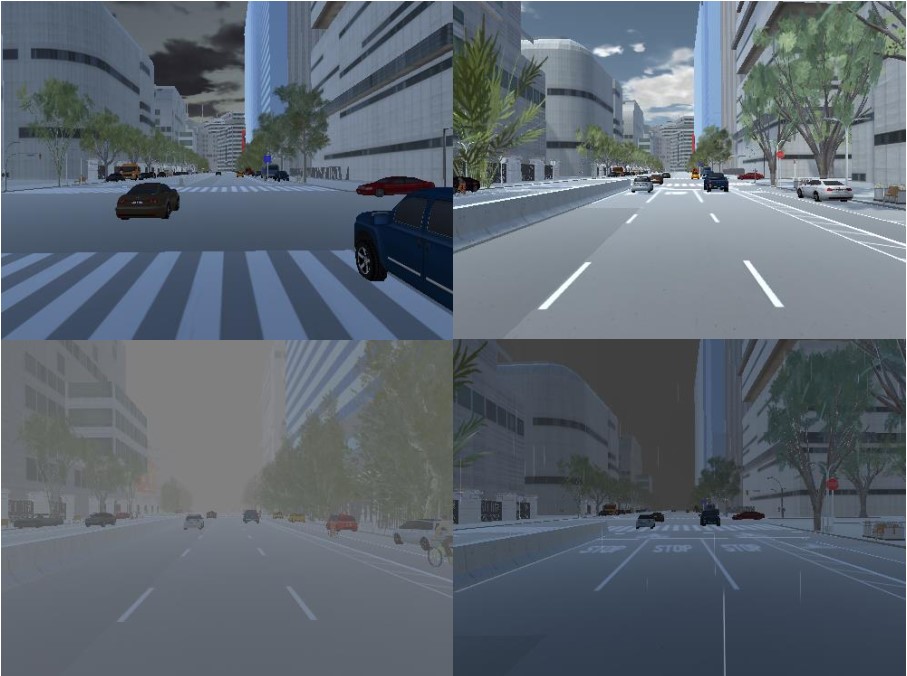}
    \caption{Diversity of illuminations and weather conditions. Top: Virtual images took at 6:00 am (left) and 12:00 pm (right). Bottom: Virtual images with weather of foggy (left) and rainy (right).}
\end{figure}
\begin{figure}
    \centering
    \includegraphics[width=88mm]{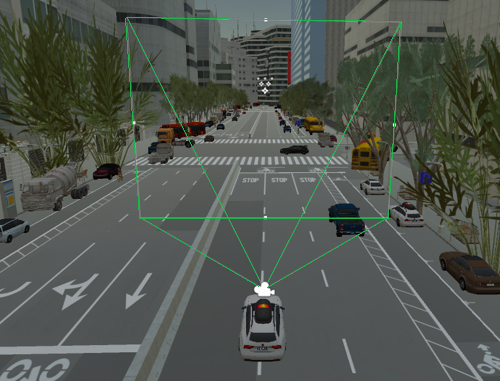}
    \caption{\textcolor{\mycolor}{Virtual camera in the artificial scene.}}
\end{figure}
\subsection{Setting Up Virtual Dataset}
In order to increase the diversity of the virtual dataset, we configured different weather (cloudy, sunny, rainy and foggy) and illumination (from sunrise to sunset) conditions for the artificial scene as shown in Fig. 5. These changes are deemed to have a significant effect on the performance of object detectors in real world \cite{wang2012measuring}\cite{wang2016multi}\cite{zhang2017advances}. We placed a virtual camera on a moving car that was used for capturing images in the scene. To produce obvious change in object's appearance, we set different parameters for the camera including height, orientation and the field of view. The virtual camera is illustrated in Fig. 6. The sight distance of the camera can be adjusted and it is much longer in practice. We placed several cars, buses and trucks on the lanes which can move following the instruction of the scripts. More vehicles were put near the roads in different manners, sparse or dense, to create diverse occlusion levels. For the purpose of testing and improving the performance of object detectors on objects with distinct colors and poses, we achieved real-time color and pose change of the interesting objects in artificial scene while the virtual camera was collecting images. Based on the above techniques, we built the virtual dataset named ParallelEye \cite{Xuan2017The}, which is composed of three sub datasets.\par
\begin{figure}
    \centering
    \includegraphics[width=88mm]{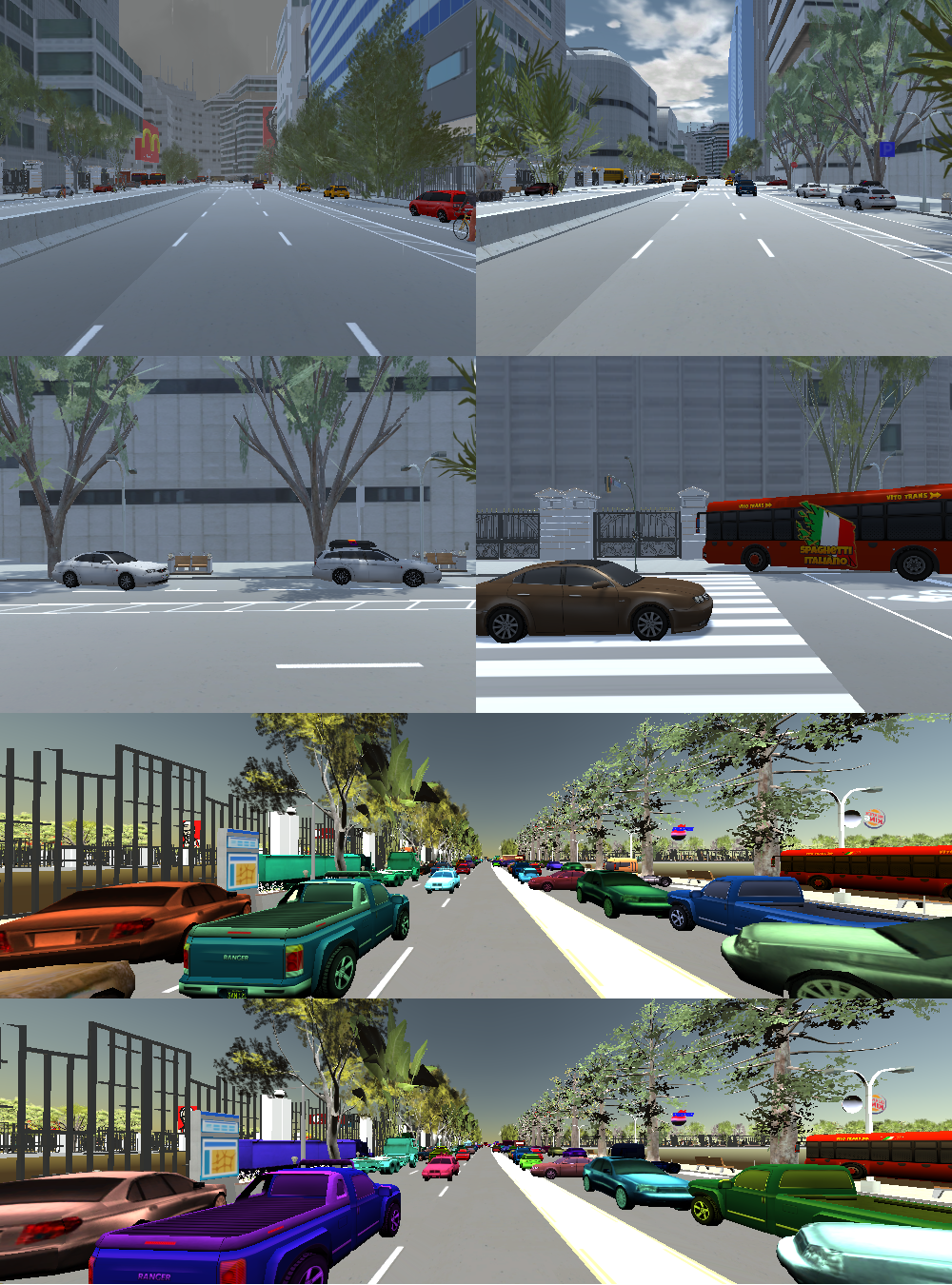}
    \caption{Sample images of three virtual sub datasets. First row: ParallelEye\_01. Second row: ParallelEye\_02. Third and fourth rows: ParallelEye\_03.}
\end{figure}
ParallelEye\_01 is the first part of the virtual dataset which was set up with an on-board camera looking at five directions (i.e., 0, $\pm$15, $\pm$30 degree with respect to the moving direction). Therefore, the camera had a long sight distance to capture small and far objects. In ParallelEye\_02, the orientation of the camera was adjusted to ±90 degree with respect to the moving direction and we set the vehicles to rotate around their axes. \textcolor{\mycolor}{Occlusion was not intentionally introduced to get a better understanding of the effect of pose change on the trained models.} ParallelEye\_03 was designed to investigate the influence of color and occluded condition on the performance of the object detector. We placed the vehicles crowdedly and changed their colors in every frame. \textcolor{\mycolor}{The camera was set to look forward.} Sample images from these three parts of the virtual dataset are shown in Fig. 7.\par
\subsection{Dataset Properties}
\begin{figure*}
    \centering
    \includegraphics[width=\textwidth]{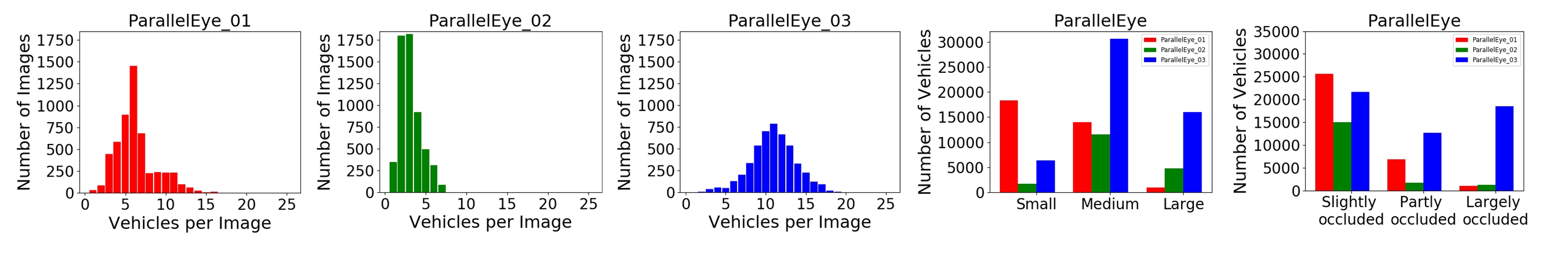}
    \caption{Object occurrence, object geometry and occlusion level of our virtual datasets. This figure shows (from left to right): Distribution of the number of instances within an image, object area distribution and occlusion level.}
\end{figure*}

ParallelEye includes 15,931 images and annotations of three classes of objects (car, bus and truck) in VOC format. The numbers of images and objects in three sub datasets are recorded in Table I. The bar graphs in Fig. 8 depict the object occurrence and object geometry statistics as well as the occlusion level of three sub datasets. The objects are labelled as ``Small" whose areas are smaller than 32$\times$32 pixels and ``Large" for those whose areas are larger than 96$\times$96 pixels. The rest are labelled as ``Medium". For occlusion level, the objects are labelled as ``Slightly occluded" whose  occlusion rates are less than 0.1 and ``Largely occluded" for those whose occlusion rates are more than 0.35 and ``Partly occluded" for the rest. It is clear to see that ParallelEye\_01 and ParallelEye\_03 are more crowded than ParallelEye\_02, which means there are less objects in one image in ParallelEye\_02 compared with the other sub datasets. More importantly, ParallelEye\_01 has more small objects and ParallelEye\_03 has a higher occlusion level.

\par
\subsection{\textcolor{\mycolor}{Discussion}}
\textcolor{\mycolor}{In the experiments, our platform can work at 8-12 fps (frames per second) to produce virtual images with different annotations on our workstation. It is efficient compared with manual labor, for example, it usually takes dozens of minutes to finish the labeling work on a single image containing various categories used for segmentation task.}\par
\textcolor{\mycolor}{This virtual dataset was mainly built for researches on intelligent vehicles. For now, we included three common types of objects. Other objects such as person and motorbikes have not yet be contained due to their less appearance from the view of the virtual camera shown in Fig. 6. In the future, we will add more objects in our virtual scene under the framework shown in Fig. 3 to build a dataset containing more categories. Specifically, add the models such as person, motorbikes and animals in the virtual scene during step \textcircled{3} and take them into account in the C\# scripts during steps \textcircled{4}\textcircled{5}\textcircled{6}. Then, we are able to capture images containing these objects with diverse annotations.}

\par
\textcolor{\mycolor}{It is also worth mentioning the ways to reduce computational complexity when we generalize the virtual scene to a bigger city or other large-scale environments by importing the corresponding maps during step \textcircled{1} and more models at step \textcircled{3} in Fig. 3. Great computational capacity is demanded as the scale of the virtual world increases if no action is taken. In practice, there are two widely-used methods to handle this problem. One is that we can only take into account the objects which are visible to the virtual camera instead of all the objects in the scene via some tricks like occlusion culling. Besides, it is also an effective way to replace the intricate textures and structures with rougher ones for the objects that are far from the camera. These measures help to decrease the workload of CPU and GPU of the platform, e.g., the number of vertex to compute, thus leading to an acceptable frame rate even when a large-scale scene is running.}
\newcolumntype{b}{>{\centering\arraybackslash}X}
\newcolumntype{s}{>{\hsize=.6\hsize}>{\centering\arraybackslash}X}
\begin{table}[htbp]
\centering
    \caption{Numbers of images and objects in three virtual sub datasets}
    \begin{tabularx}{88mm}{bssss}
    \toprule
    Sub dataset & Number  of images & Number of  cars & Number of buses & Number of trucks\\ \midrule
    ParallelEye\_01 & 5,313 & 13,653 & 8,593 & 11,132\\
    ParallelEye\_02 & 5,781 & 7,892  & 6,881 & 3,281\\
    ParallelEye\_03 & 4,837 & 31,369 & 16,863& 4,697\\ \bottomrule
    \end{tabularx}
\end{table}

\section{Experiments}
In our experiments, all the images and annotations were stored in the form of PASCAL VOC. For Faster R-CNN, the weights were initialized with ImageNet pre-trained weights. \textcolor{\mycolor}{For the experiments of DPM, we used a three-component model for each category and the ratio of positive and negative examples was set as 1:3.}

\begin{figure*}
    \centering
    \includegraphics[width=0.85\textwidth]{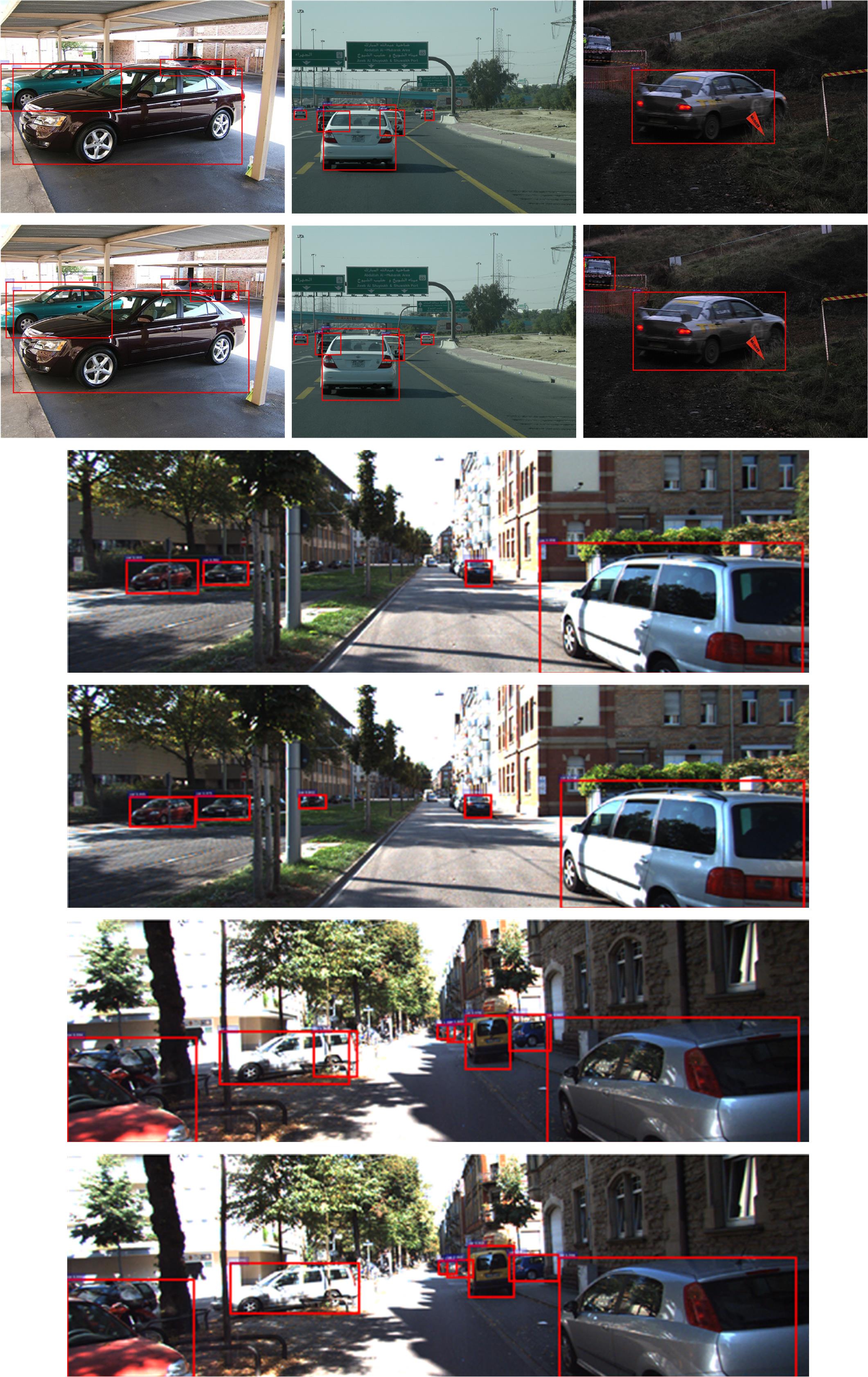}
    \caption{Examples of object detection results (VGG-16). The upper one of every couple: Objects detected by the model trained purely with the real training set. The lower one of every couple: Objects detected by the model trained with real and virtual datasets.  }
\end{figure*}

\subsection{Training Experiments on VOC+COCO and ParallelEye}

To obtain a real dataset containing three common traffic objects, i.e., car, bus and truck, we selected 1,997 images including car and bus from PASCAL VOC \cite{everingham2010pascal} and \textcolor{\mycolor}{1,000 images} containing truck from MS COCO \cite{lin2014microsoft} that were transformed to VOC style. \textcolor{\mycolor}{MS COCO includes more images containing truck. But we only chose 1,000 images and excluded images where truck shares a fairly small proportion.} These images and corresponding annotations from VOC and COCO were combined together and randomly divided into training set and testing set with the proportion of 3:1. Firstly, we trained the Faster R-CNN model with the real training set. Initial learning rate was set to 0.001 and decreased by the factor of 10 \textcolor{\mycolor}{after 50,000 iterations.} We chose one image as the batch size and momentum factor of 0.9 with the weight decay factor of 0.0005. Then, we randomly picked 2,000 images out from the virtual dataset ParallelEye and combined them with the real training data as a mixed training set to train another model with the same setting. \textcolor{\mycolor}{We used VGG-16 and ResNet-50 for Faster R-CNN respectively in our experiments. Finally, we carried out experiments for DPM using the same datasets.}

\par
These models were evaluated on the real testing set generated from PASCAL VOC and MS COCO. We followed the steps of the standard evaluation procedure of PASCAL VOC to calculate the Average Precision (AP) of each category. \textcolor{\mycolor}{The intersection over union (IoU) was set to 0.7 for Faster R-CNN and 0.5 for DPM.} The results are shown in Table II. The examples of detection results for Faster R-CNN based on VGG-16 architecture are shown in Fig. 9. \textcolor{\mycolor}{We noticed that the AP on bus for DPM is higher than Faster R-CNN. This may be caused by the fact that the shape of bus is not as flexible as car and truck, so it is easier to learn. Also, the number of bus is less than the other two types of objects \cite{everingham2010pascal} which is adverse for Faster R-CNN. Because deep learning models usually require more data to learn than the traditional ones. We also noticed that after we introducing the truck category from MS COCO, the AP of car and bus is decreased compared with models trained purely on the PASCAL VOC dataset for Faster R-CNN \cite{ren2015faster}, which may be interpreted by the fact that MS COCO dataset is more challenging for object detection and contains more difficult images of objects that are partially occluded, amid clutter, etc \cite{lin2014microsoft}.}
\begin{table}[htbp]
\centering
    \caption{Performance of models evaluated on VOC+COCO}
    \begin{tabularx}{88mm}{bssss}
    \toprule
    Model & Training dataset & Car & Bus & Truck\\ \midrule
    \multirow {2}{2cm}{\textcolor{\mycolor}{DPM}} & \textcolor{\mycolor}{Real} & \textcolor{\mycolor}{0.482} & \textcolor{\mycolor}{0.715} & \textcolor{\mycolor}{0.482}\\
    & \textcolor{\mycolor}{Mixed} & \textbf{\textcolor{\mycolor}{0.496}} &\textbf{\textcolor{\mycolor}{0.721}} & \textbf{\textcolor{\mycolor}{0.499}}\\ \midrule
    \multirow {2}{2cm}{\textcolor{\mycolor}{Faster R-CNN (VGG-16)}} & \textcolor{\mycolor}{Real} & \textcolor{\mycolor}{0.736} & \textcolor{\mycolor}{0.639} & \textcolor{\mycolor}{0.642}\\
    & \textcolor{\mycolor}{Mixed} & \textbf{\textcolor{\mycolor}{0.750}} & \textbf{\textcolor{\mycolor}{0.644}} & \textbf{\textcolor{\mycolor}{0.657}}\\ \midrule
     \multirow {2}{2cm}{\textcolor{\mycolor}{Faster R-CNN (ResNet-50)}} & \textcolor{\mycolor}{Real} & \textcolor{\mycolor}{0.730} & \textcolor{\mycolor}{0.616} & \textcolor{\mycolor}{0.663}\\
    & \textcolor{\mycolor}{Mixed} & \textbf{\textcolor{\mycolor}{0.754}} & \textbf{\textcolor{\mycolor}{0.631}} & \textbf{\textcolor{\mycolor}{0.679}}\\ \bottomrule
    \end{tabularx}
\end{table}

\subsection{Training Experiment on KITTI and ParallelEye}
6,684 images with annotations containing the car object were picked out from the KITTI dataset and divided into real training set and testing set at the ratio of one to one. First, we trained the Faster R-CNN detector purely with the real training set. Next, 4,000 images with annotations containing the car object were randomly selected from ParallelEye dataset that were used for pre-training the Faster R-CNN model. Then, we fine-tuned the pre-trained model using the real training set. \textcolor{\mycolor}{And we performed experiments on DPM with real and mixed data.} These experiments were executed with the same setting as experiments on VOC+COCO. These trained models were tested on the real KITTI testing set and the Average Precision is recorded in Table III. Fig. 9 depicts the examples of the detection results for Faster R-CNN based on VGG-16 architecture.
\begin{table}[htbp]
\centering
    \caption{Performance of models evaluated on KITTI }
    \begin{tabular}{C{2cm}C{3.7cm}C{1.8cm}}
    \toprule
    Model & Training dataset & Test on KITTI \\ \midrule
    \multirow {2}{2cm}{\textcolor{\mycolor}{DPM}} & \textcolor{\mycolor}{KITTI} & \textcolor{\mycolor}{0.516}\\
    & \textcolor{\mycolor}{Mixed} & \textbf{\textcolor{\mycolor}{0.527}}\\ \midrule
    \multirow {2}{2cm}{Faster R-CNN (VGG-16)} & KITTI & 0.741\\
    &ParallelEye $\rightarrow$ KITTI & \textbf{0.757}\\ \midrule
    \multirow {2}{2cm}{\textcolor{\mycolor}{Faster R-CNN (ResNet-50)}} & \textcolor{\mycolor}{KITTI} & \textcolor{\mycolor}{0.785}\\
    & \textcolor{\mycolor}{ParallelEye $\rightarrow$ KITTI} & \textbf{\textcolor{\mycolor}{0.798}}\\ \bottomrule
    \end{tabular}
\end{table}
\subsection{Testing Experiments on KITTI and ParallelEye}
In this section, we investigate the potential of using our virtual dataset to test the object detector trained on real dataset. KITTI was chosen as the real dataset to train Faster R-CNN models due to its diversity of object area and occlusion condition \cite{geiger2012we}. 6,684 images with annotations containing the car object were picked out from the KITTI dataset and divided into training set and testing set at the ratio of one to one. We named the full training set Train\_Full to avoid confuse. The Train\_Full set possessed 3,342 images with 28,742 cars. First, we trained a Faster R-CNN detector with Train\_Full. This detector was used as the reference for the detectors below. To evaluate the impact of small objects on model's performance, we deleted these objects in Train\_Full whose areas were smaller than 3,600 pixels according to the annotations. By this standard, we got 2,419 images containing 14,346 cars. We called the devised training set Train\_Large which was used to train the second Faster R-CNN detector. The third detector was trained with so-called Train\_Visible which only kept objects labeled as "fully visible" in Train\_Full and had 3,113 images with 13,457 cars. All the experiments were carried out with the same setting as those in VOC+COCO.\par
\begin{table}[htbp]
\centering
    \caption{Evaluation with purposefully designed virtual datasets }
    \begin{tabular}{C{1.5cm}C{2cm}C{2cm}C{2cm}}
    \toprule
    Training dataset & Test on ParallelEye\_01 & \textcolor{\mycolor}{Test on ParallelEye\_02} & Test on ParallelEye\_03 \\ \midrule
    Train\_Full & \textbf{0.485} & \textbf{\textcolor{\mycolor}{0.570}} & \textbf{0.585} \\
    Train\_Large & 0.256 & \textcolor{\mycolor}{0.508} & 0.467 \\
    Train\_Visible & 0.348 & \textcolor{\mycolor}{0.433} & 0.396 \\ \bottomrule
    \end{tabular}
\end{table}
These three detectors were tested on ParallelEye\_01 characterized by small area of objects on average and ParallelEye\_03 marked by high level of occlusion as well as ParallelEye\_02 with larger objects and lower occlusion level. We calculated the Average Precision in the manner of PASCAL VOC. The results are recorded in Table IV. For the purpose of making the results more explicit, we also calculated the rate of descent of AP after we removed small objects and occluded objects respectively from the training set. The AP of model trained with Train\_Full set was regarded as the reference and the results are shown in Table V.

\begin{table}[htbp]
\centering
    \caption{Rate of descent on virtual datasets }
    \begin{tabularx}{88mm}{C{1.5cm}C{2cm}C{2cm}C{2cm}}
    \toprule
    Training dataset & Rate of descent on ParallelEye\_01 & \textcolor{\mycolor}{Rate of descent on ParallelEye\_02} & Rate of descent on ParallelEye\_03\\ \midrule
    Train\_Large & \textbf{47.2\%} & \textcolor{\mycolor}{10.9\%} & 20.2\%\\
    Train\_Visible &28.2\% & \textbf{\textcolor{\mycolor}{24.0\%}} & \textbf{32.3\%} \\ \bottomrule
    \end{tabularx}
\end{table}

\subsection{Discussion}

On the one hand, the results above show that our virtual datasets are viable to improve the performance of object detector when used for training it together with the real dataset. On the other hand, we can conclude that our purposefully designed virtual datasets are potential tools to assess the performances of trained models from a specific aspect. The results in Table IV and Table V show that with small objects removed from the training set, performance of the model became worse on all sub datasets while a bigger rate of descent of AP occurred on ParallelEye\_01 in the testing phase, which may result from the smaller average area of object of ParallelEye\_01. And ParallelEye\_03 witnessed a huger drop of the rate of AP after we deleted occluded objects from the training set because ParallelEye\_03 has a higher occlusion rate.

\section{Conclusion}
This paper presents a pipeline to build artificial scenes and virtual datasets possessing some specific characteristics we desire like the occlusion level, area of objects and so on under the framework of Parallel Vision. We prove that mixing the virtual dataset and several real datasets to train the object detector helps to improve the performance. Also, we investigate the potential of testing the trained models on a specific aspect using intentionally designed virtual datasets. This work may help deep learning researchers to get a better understanding of their models especially in the areas of autonomous driving.

\bibliographystyle{IEEEtran}
\bibliography{IEEEabrv,mine}
%


\begin{IEEEbiography}[{\includegraphics[width=1in,height=1.25in]{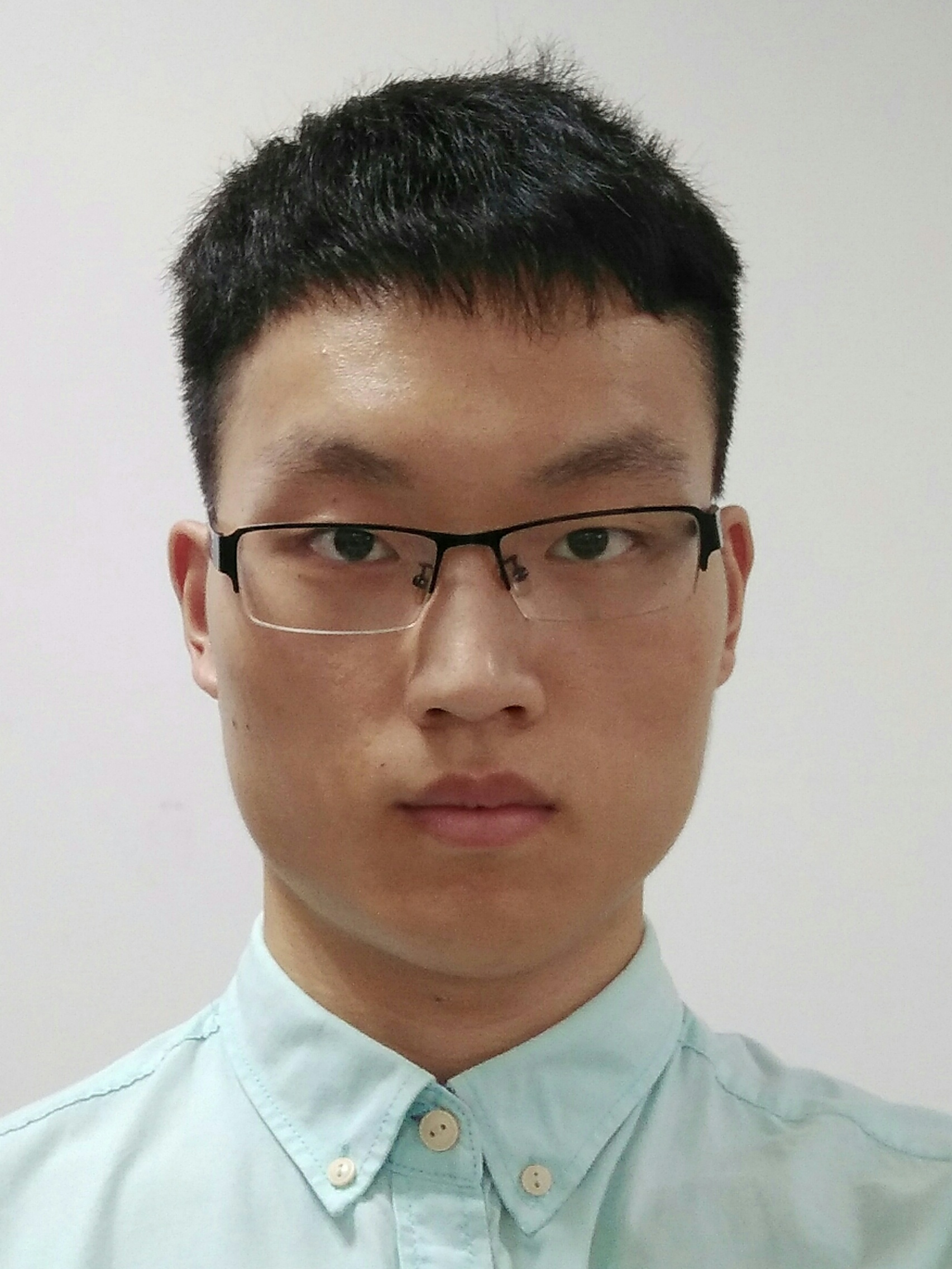}}]{Yonglin Tian}
received his bachelor degree from the University of Science and Technology of China in 2017. He is currently a Master student at the Department of Automation, University of Science and Technology of China as well as the State Key Laboratory of Management and Control for Complex Systems, Institute of Automation, Chinese Academy of Sciences. His research interests include computer vision and pattern recognition.
\end{IEEEbiography}

\begin{IEEEbiography}[{\includegraphics[width=1in,height=1.25in]{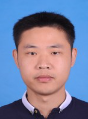}}]{Xuan Li}
received the M.S. degree from the Changsha University of Science and Technology, Changsha, China, in 2013. He is currently working toward the Ph.D. degree in control theory and control engineering with the Key Laboratory of Intelligent Control and Decision of Complex Systems, Beijing Institute of Technology, Beijing, China. His research interests include computer vision, pattern recognition, intelligent transportation systems, and 3D printing.\end{IEEEbiography}

\begin{IEEEbiography}[{\includegraphics[width=1in,height=1.25in]{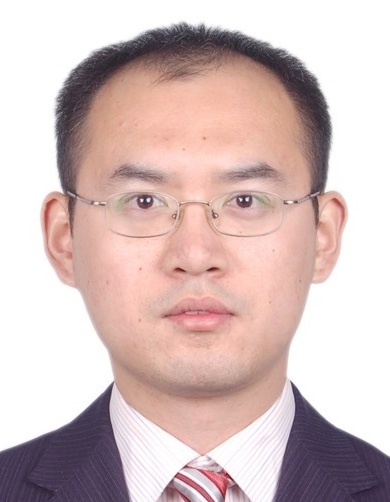}}]{Kunfeng Wang}
(M'17) received his Ph.D. in Control Theory and Control Engineering from the Graduate University of Chinese Academy of Sciences, Beijing, China, in 2008. He joined the Institute of Automation, Chinese Academy of Sciences and became an Associate Professor at the State Key Laboratory of Management and Control for Complex Systems. From December 2015 to January 2017, he was a Visiting Scholar at the School of Interactive Computing, Georgia Institute of Technology, Atlanta, GA, USA. His research interests include intelligent transportation systems, intelligent vision computing, and machine learning. He is a Member of IEEE.
\end{IEEEbiography}

\begin{IEEEbiography}[{\includegraphics[width=1in,height=1.25in]{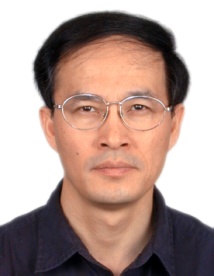}}]{Fei-Yue Wang}
(S'87-M'89-SM'94-F'03) received his Ph.D. in Computer and Systems Engineering from Rensselaer Polytechnic Institute, Troy, New York in 1990. He joined the University of Arizona in 1990 and became a Professor and Director of the Robotics and Automation Lab (RAL) and Program in Advanced Research for Complex Systems (PARCS). In 1999, he founded the Intelligent Control and Systems Engineering Center at the Institute of Automation, Chinese Academy of Sciences (CAS), Beijing, China, under the support of the Outstanding Oversea Chinese Talents Program from the State Planning Council and ``100 Talent Program" from CAS, and in 2002, was appointed as the Director of the Key Lab of Complex Systems and Intelligence Science, CAS. In 2011, he became the State Specially Appointed Expert and the Director of The State Key Laboratory of Management and Control for Complex Systems. Dr. Wang's current research focuses on methods and applications for parallel systems, social computing, and knowledge automation. He was the Founding Editor-in-Chief of the International Journal of Intelligent Control and Systems (1995-2000), Founding EiC of IEEE ITS Magazine (2006-2007), EiC of IEEE Intelligent Systems (2009-2012), and EiC of IEEE Transactions on ITS (2009-2016). Currently he is EiC of China's Journal of Command and Control. Since 1997, he has served as General or Program Chair of more than 20 IEEE, INFORMS, ACM, ASME conferences. He was the President of IEEE ITS Society (2005-2007), Chinese Association for Science and Technology (CAST, USA) in 2005, the American Zhu Kezhen Education Foundation (2007-2008), and the Vice President of the ACM China Council (2010-2011). Since 2008, he is the Vice President and Secretary General of Chinese Association of Automation. Dr. Wang is elected Fellow of IEEE, INCOSE, IFAC, ASME, and AAAS. In 2007, he received the 2nd Class National Prize in Natural Sciences of China and awarded the Outstanding Scientist by ACM for his work in intelligent control and social computing. He received IEEE ITS Outstanding Application and Research Awards in 2009 and 2011, and IEEE SMC Norbert Wiener Award in 2014.
\end{IEEEbiography}

\end{document}